%% file: anonymous-submission-latex-2025.tex
\documentclass[letterpaper]{article} % DO NOT CHANGE THIS
\usepackage{aaai25}  % DO NOT CHANGE THIS
\usepackage{times}  % DO NOT CHANGE THIS
\usepackage{helvet}  % DO NOT CHANGE THIS
\usepackage{courier}  % DO NOT CHANGE THIS
\usepackage[hyphens]{url}  % DO NOT CHANGE THIS
\usepackage{graphicx} % DO NOT CHANGE THIS
\usepackage{natbib}  % DO NOT CHANGE THIS AND DO NOT ADD ANY OPTIONS TO IT
\usepackage{caption} % DO NOT CHANGE THIS AND DO NOT ADD ANY OPTIONS TO IT
\usepackage{algorithm}
\usepackage{algorithmic}
\usepackage{tikz}
\usetikzlibrary{positioning}
\usepackage{newfloat}
\usepackage{listings}
\DeclareCaptionStyle{ruled}{labelfont=normalfont,labelsep=colon,strut=off} % DO NOT CHANGE THIS
\floatstyle{ruled}
\newfloat{listing}{tb}{lst}{}
\floatname{listing}{Listing}
\title{Handling Uncertainty in Health Data using Generative Algorithms}
\author {
    Mahdi Arab Loodaricheh\textsuperscript{\rm 1},
    Neh Majmudar\textsuperscript{\rm 1},
    Anita Raja\textsuperscript{\rm 1,2}
    Ansaf Salleb-Aouissi\textsuperscript{\rm 3}
}
\affiliations {
    \textsuperscript{\rm 1}City University of New York, Graduate Center\\
    \textsuperscript{\rm 2}City University of New York, Hunter College\\
    \textsuperscript{\rm 3}Columbia University\\
    marabloodaricheh@gradcenter.cuny.edu, neh.majmudar@gradcenter.cuny.edu, anita.raja@hunter.cuny.edu, ansafsalleb@columbia.edu
}

\usepackage{bibentry}
\begin{document}

\maketitle

\begin{abstract}

Understanding and managing uncertainty is crucial in machine learning, especially in high-stakes domains like healthcare, where class imbalance can impact predictions. This paper introduces RIGA, a novel pipeline that mitigates class imbalance using generative AI. By converting tabular healthcare data into images, RIGA leverages models like cGAN, VQVAE, and VQGAN to generate balanced samples, improving classification performance. These representations are processed by CNNs and later transformed back into tabular format for seamless integration. This approach enhances traditional classifiers like XGBoost, improves Bayesian structure learning, and strengthens ML model robustness by generating realistic synthetic data for underrepresented classes.

\end{abstract}

% Uncomment the following to link to your code, datasets, an extended version or similar.
%
% \begin{links}
%     \link{Code}{https://aaai.org/example/code}
%     \link{Datasets}{https://aaai.org/example/datasets}
%     \link{Extended version}{https://aaai.org/example/extended-version}
% \end{links}

\input{intro_new}
\input{related}

\input{approach}

\input{experiments}

\input{results}

\input{summary}
\input{conclusions}
\input{ack}

\bibliography{aaai25.bib}

\appendix

\input{appendix}

\end{document}

%% file: intro_new.tex
\section{Introduction}
\label{sec:intro}

Decision-making in complex clinical settings often involves high-stakes choices regarding patient care under conditions of incomplete information. Numerous clinical decision support (CDS) models have been proposed to assist clinicians in navigating these scenarios. Bayesian Networks (BNs) \cite{Koller2009} represent a particularly promising approach, as they meet essential requirements for CDS in complex, real-world datasets: (1) uncovering novel patterns and (2) enhancing explainability. The capability of BNs to visually represent probabilistic dependencies between features makes them especially beneficial for clinicians who need to understand intricate feature interactions when complete relationships are not well established. BNs learned in an unsupervised manner provide one of the few modeling approaches capable of simultaneously addressing these dual objectives.

Furthermore, class imbalance is a pervasive challenge in health datasets, where an uneven distribution of outcome classes complicates the application of machine learning techniques. For instance, we seek to determine the risk of adverse pregnancy outcomes, specifically preterm birth (PTB) \cite{purisch2017epidemiology} among nulliparous women (i.e., those with no prior births). PTB, defined as delivery before 37 weeks of gestation, represents a minority outcome, affecting approximately 10\% of pregnancies in the Nulliparous Pregnancy Outcomes Study (nuMoM2b) dataset \cite{Haas2015nuMoM2b}, which comprises observational data from over 10,000 pregnancies, encompassing more than 4,600 features tracked across four prenatal visits. The inherent class imbalance necessitates sophisticated modeling approaches to accurately identify risk factors and enhance prediction efficacy. Although a prior history of PTB is a predominant risk factor for subsequent preterm deliveries, a substantial number of PTBs occur in nulliparous women with no previous childbirth history \cite{koullali2016risk}, and many occur without discernible symptoms or established clinical predictors \cite{iams2001preterm, huang2018investigation, manuck2017racial}.

This paper introduces RIGA (Robustness using Imbalance-Resilient Generative Augmentation), a four-phase approach to mitigate class imbalance via data augmentation~\citep{mumuni2022data, maharana2022review, mikolajczyk2018data, nanni2021comparison}. Generative modeling, particularly Generative Adversarial Networks (GANs)~\citep{goodfellow2014generative}, is a notable method for creating synthetic data that resembles the original dataset, proving highly effective in data augmentation tasks. GANs excel with image data~\citep{aggarwal2018neural}, thus transforming the data into image representations~\citep{sharma2019deepinsight, Zhu2021} before augmentation is advantageous.

The core idea of this work is to transform non-image data into image-like formats for input into a Convolutional Neural Network (CNN). Methods like similarity measurement and dimensionality reduction (e.g., t-SNE, kPCA) help create a 2D representation capturing feature relationships. The nuMoM2b dataset exhibits a complex hierarchical structure, with both temporal and non-temporal feature dependencies. By converting such tabular data into images, generative models can leverage hidden structures, leading to more realistic and useful samples.

The {\bf key contributions} of this work include using a data-to-image conversion method to transform rows of the nuMoM2b dataset into images, providing researchers with a novel perspective on data structure. Next, generative models are employed to augment these transformed medical datasets. Following augmentation, a lossless inverse transformation restores the data to its original form, enabling the use of traditional classification algorithms. Finally, a Bayesian Network pipeline is applied to evaluate the augmentation's impact on determining feature relationships within the nuMoM2b dataset.

%% file: related.tex
\section{Related Work}
\label{sec:related}

% {\bf Augmentation of Tabular Data}: Synthetic Minority Oversampling Technique (SMOTE) \citep{chawla2002smote} is one of the most popular data augmentation methods for tabular data. SMOTE uses a nearest neighbor algorithm to generate new minority class instances by slightly perturbing existing minority class data. While this method often works well, it is not suitable for datasets like nuMoM2b, where the minority class is comprised of a complex and heterogeneous population. Without understanding the underlying structure of the minority population, it is not possible for SMOTE to produce quality synthetic samples. Adaptive Synthetic (ADASYN) \citep{adasyn} sampling is a method similar to SMOTE, which generates synthetic samples in proportion to the difficulty of the classification for a given minority class. ADASYN faces similar issues to SMOTE in the context of complex minority populations.

{\bf Augmentation of Tabular Data}: Synthetic Minority Oversampling Technique (SMOTE) \citep{chawla2002smote} is a widely-used data augmentation method for tabular data, generating new minority class instances using a nearest neighbor approach. However, SMOTE struggles with complex, heterogeneous datasets like nuMoM2b, where understanding the underlying structure of the minority class is crucial. Adaptive Synthetic (ADASYN) \citep{adasyn}, which generates synthetic samples based on the difficulty of classification, also faces similar limitations in such contexts.

\begin{figure*}[h]
  \centering
  \includegraphics[width=1.0\textwidth]{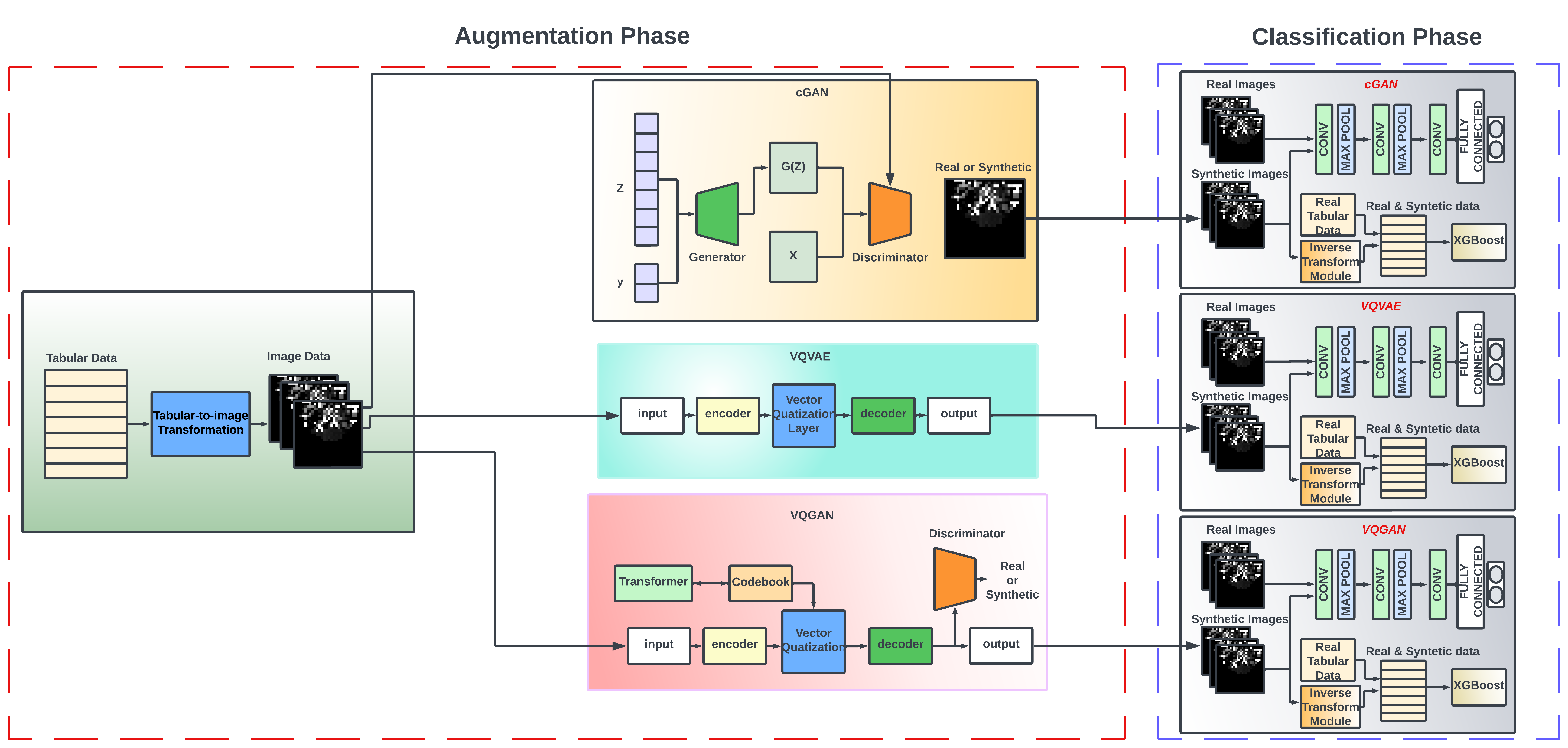}
  %\caption{The pipeline of Classification approach}
   \caption{RIGA Pipeline of Augmentation and Classification Method: Left - Tabular-to-Image Conversion with cGAN, VQVAE, and VQGAN Training. (X: Data sample, y: Condition , and Z: Latent Noise Vector) Right - Dual Classification Paths: Top - Real and Generated Images Classified via CNN; Bottom - Synthetic Images Converted Back to Tabular and Classified with XGBoost.} 
  \label{fig:method_fig}
\end{figure*}

% Attempts have been made to augment tabular data using deep neural networks, but are not widely popular due to the difficulties tabular data poses for neural learners, such as mixed numerical and categorical data types. Conditional Tabular GAN (CTGAN) \citep{ctgan} is a Conditional GAN architecture adapted to tabular data. CTGAN tackles mixed continuous and discrete features by adapting the normalization method to the mode of each feature. This method was found to outperform other GAN-based approaches for augmenting tabular data. We show that our method performs better than CTGAN on the datasets we test, suggesting that conversion to images may result in better synthetic data. 
Deep neural networks have been used to augment tabular data, but their popularity is limited due to challenges with mixed numerical and categorical features. 
% \anita{Narrative flow seems off - why bring up CTGAN here when generative models are introduced in next paragraph. Move it later}
Conditional Tabular GAN (CTGAN) \citep{ctgan} adapts to tabular data by normalizing mixed features and has been found to outperform other GAN-based methods. However, our approach outperforms CTGAN, suggesting that converting tabular data to images can produce better synthetic data. 
% \mahdi{ctgan is an augmentation method for tabular data. change the next title to Generative models for images}

% \anita{if you use the word 
% may" you should explain here in one sentence and in your results where other approaches are better and where CTGAN is better}  

% \textcolor{red}{remove FAKE use "synthetic"}
% \anita{yes in text and in image}
% \subsection{Generative models}

% GANs \citep{goodfellow2014generative} are a class of deep neural architectures that have demonstrated marked success generating synthetic data, most notably images. GANs are comprised of two competing networks, a generator and a discriminator, which are trained in tandem. The generator produces novel data instances and while the discriminator attempts to discern which instances are genuine (coming from the original dataset), and which are synthetic. As the discriminator improves, the generator must learn to create increasingly better synthetic examples. Due to their success in producing synthetic images that match a target distribution, GANs are one of the most popular methods employed in the augmentation of image datasets \citep{image_aug_survey}. In this setting, the Conditional GAN (cGAN) \citep{cgan} architectural variant is common, as it is sensitive to data imbalances--this is the GAN variant we use in our work.
{\bf Generative models for Images}: GANs \citep{goodfellow2014generative} are deep neural architectures successful in generating synthetic data, particularly images. GANs consist of a generator and a discriminator trained in tandem—the generator creates novel data while the discriminator distinguishes real from synthetic instances. This competition improves both networks over time. GANs are widely used for augmenting image datasets \citep{image_aug_survey}. In our work, we utilized three distinct generative models to compare their performance and evaluate their effectiveness across different underlying data structures. Conditional GAN (cGAN) \citep{cgan} variant effectively handles data imbalances. 
VQVAE (Vector Quantized Variational Autoencoder)~\citep{vandenoord2017neural} learns discrete latent representations for high-quality reconstructions, reducing input dimensionality. The VQ process involves finding the nearest codebook entry to a given pixel or region. The quantization is represented as \( q(z) = \arg\min_{c \in C} \|z - c\|^2 \), where \( C \) is the codebook.
VQVAE reduces blurry outputs compared to traditional VAEs, but generated quality depends on codebook size. VQGAN~\citep{esser2021taming} combines VQVAE with GANs to improve training efficiency and image quality. However, it requires careful tuning due to challenges like mode collapse.

% \subsection{Tabular to image transformation}

{\bf Tabular to image transformation}: \textit{DeepInsight} \citep{sharma2019deepinsight} converts non-image data into images by recovering local structure with a similarity-based method like KNN and mapping features to pixels based on similarity. Pixel locations correspond to features, and pixel intensities represent feature values, allowing CNN application to non-image data. 
% \anita{for IJCAI 2024 January submission, the goal will be to find holes if any on the previous 2 sentences and come up with your own new and better method}
A limitation is that many pixels do not represent features \citep{Zhu2021}, but our approach addresses this by using an appropriate image size with one pixel per feature, making the transformation invertible. We utilize this to augment data in image form and convert it back to tabular format for traditional classification. \citep{ANDRESINI2021108} introduced MAGNETO, adapting \textit{DeepInsight} and GANs for intrusion detection in traffic data using a CNN, but did not explore inverse transformation to tabular format for use with traditional models, which could have allowed for a direct comparison.

%% file: approach.tex
\section{Approach: The RIGA pipeline}

% The RIGA pipeline has three stages: tabular data is converted into images, generative models create enhanced datasets, and classifiers handle image and non-image data for effective classification.

\subsection{Phase 1: Data Transformation}

During the initial stage, we apply the {\it DeepInsight} method \citep{sharma2019deepinsight} to convert our tabular data into images. This transformation allows us to subsequently leverage generative models. Each row  (feature vector) within our dataset undergoes a conversion process, resulting in images of dimensions 28x28. The  feature value is mapped to the feature's location in the image which in turn is determined by its similarity. The data matrix, initially consisting of n rows and d features, is transposed to have d rows and n columns. Subsequently, this transposed matrix undergoes dimensionality reduction techniques such as t-SNE, pinpointing each feature's location in 2D space to derive a 2D plane. As a result, every row in our dataset is transformed into an image.
Using this method, along with ensuring the correct image size (28x28 in our case), guarantees that each feature is assigned a specific location in the 2D space, allowing for a fully invertible transformation without any loss of information.

\subsection{Phase 2: Augmentation}

In the second phase, we implement three different generative models to generate additional data for the underrepresented class in our dataset. This approach effectively expands the dataset by introducing synthetic data points, helping to address class imbalance. During this phase, we conditionally train images using these models, based on the distinctions between the two classes in our dataset. Once trained, we use the generator components of each model to produce synthetic images for the class with fewer samples. The augmentation phase workflow in our method is illustrated in Figure~\ref{fig:method_fig}.

%  \textbf{cGAN}:In this model, we input the real dataset along with binary labels that serve as a condition for both the generator and discriminator. During training, the generator learns to create images that match the given labels, and at the end of training, it can generate new images conditioned on either of the two labels.

% \textbf{VQ-VAE}: We train an encoder-decoder architecture to learn discrete latent representations of the dataset. After training, we use the decoder to generate synthetic images for the minority class by sampling from the learned latent space.

% \textbf{VQ-GAN}: This model uses an encoder-decoder setup combined with adversarial training. The encoder learns to compress images into discrete latent codes, and the generator reconstructs images from these codes. After training, the generator produces high-quality synthetic images for the minority class, conditioned on the learned latent codes.

\textbf{cGAN} conditions the generator and discriminator on binary labels to create images matching the labels and generates new images after training. \textbf{VQ-VAE} uses an encoder-decoder architecture to learn discrete latent representations, enabling the decoder to generate synthetic images for the minority class by sampling from the latent space. \textbf{VQ-GAN} combines an encoder-decoder setup with adversarial training, compressing images into discrete latent codes and reconstructing them to produce high-quality synthetic images for the minority class.

\subsection{Phase 3: Classification}

We leverage two distinct classification strategies: One is  {\it Direct Image Classification} which is a straightforward technique that employs standard image classification techniques on both generated and real images to make predictions. The second is {\it Leveraging Tabular Space} which can exploit {\it DeepInsight}'s  inverse transformation function to "decompose" augmented images back into tabular format. This allows application of powerful traditional machine learning algorithms on the familiar tabular data, even if they cannot handle image inputs directly, thereby unlocking the potential of superior performance from these algorithms.

In the first strategy, we employ CNNs to perform image classification. All real and synthetic images are input into the CNN model, and it undergoes training to make accurate classifications. In the second approach, we start by changing the synthetic images back into tabular data. Then, we use machine learning methods like XGBoost to analyze and classify this tabular data. The classification phase workflow in our approach is illustrated in Figure \ref{fig:method_fig}.

We convert tabular data into images where each feature has a dedicated pixel using the inverse transformation function in the {\it DeepInsight} method taking advantage of the following: 1. {\it Pixel-Perfect Mapping:} With an image size exceeding the number of features, each feature directly aligns with a unique pixel, preventing any aggregation or information loss. 2. {\it Lossless Recovery:} DeepInsight's inverse function helps us retrieve the original value of each feature using the pixel value it corresponds to after transformation. {\it 3.Faithful Reconstruction:} In the final step, this allows for a perfect reconstruction of the original data row, preserving all information without compromise.

\subsection{Phase 4: Bayesian Network Learning}

The fourth phase builds on the Unsupervised Learning under Uncertainty (U2) Pipeline, based on the methodology in~\cite{Mallia2023}. In real-world datasets like nuMoM2b, uncertainty arises from noisy data and class imbalance, obscuring meaningful patterns. High-quality synthetic data improves data-driven discovery, including feature correlation. The U2 pipeline emphasizes (a) Uncertainty: noisy data and imbalance contribute to uncertainty, and (b) Unsupervised learning: models adjust to target outcomes with minimal expert input. The pipeline includes data transformation, augmentation, classification, discretization, visualization, structure learning, and analysis. Structure learning uses the Bayesian Information Criterion (BIC), calculated as $BIC = \log P(D|G) + \frac{d}{2}\log(N)$, to construct graphical models representing data distribution. Here, $D$ is the data, $G$ is the structure, $d$ is the number of free parameters, and $N$ is the dataset size. Tabu search addresses local optima, enhancing the likelihood of finding a global optimum. The learned graphical model maps the total distribution, supporting prediction and analysis.

%% file: experiments.tex
\section{Experimental Setup}
\label{sec:experiments}

Our proposed pipeline, {\it RIGA}, tackles the problem of class imbalance in tabular datasets to enhance Bayesian structure learning and improve prediction results. In this and the following section, we empirically evaluate the predictive performance of  {\it RIGA} as compared to state-of-the-art approaches in the context of both simulated and real datasets and discuss the strengths and limitations of our approach under various conditions. This section describes the experimental setup while the next section describes the experiments and the results.

\subsection{Datasets}

% The first dataset is nuMoM2b~\citep{Haas2015nuMoM2b}, a real-world medical dataset that initially sparked our interest due to its class imbalance. The second dataset is Madelon ~\citep{guyon2006feature}, a synthetic dataset initially balanced but modified to introduce imbalance for our evaluation, allowing us to test our method in non-real-world scenarios. The third is Myocardial Infarction~\citep{misc_myocardial_infarction_complications_579}, another real-world dataset with a significant class imbalance, posing an additional challenge for our method. The forth dataset is DARWIN dataset (Diagnosis AlzheimeR WIth haNdwriting)~\citep{cilia2022darwin}  which is designed for early detection of Alzheimer's disease. 
The first dataset, nuMoM2b~\citep{Haas2015nuMoM2b}, is a real-world medical dataset that initially sparked our interest due to its class imbalance. The second, Madelon~\citep{guyon2006feature}, is a synthetic dataset modified to introduce imbalance for testing in non-real-world scenarios. The third, Myocardial Infarction~\citep{misc_myocardial_infarction_complications_579}, is a real-world dataset with significant class imbalance. The fourth, DARWIN~\citep{cilia2022darwin}, is designed for early detection of Alzheimer's disease.

\subsubsection{nuMom2b} 
The Nulliparous Pregnancy Outcomes Study: Monitoring Mothers-to-be (nuMoM2b) dataset, introduced in \citep{Haas2015nuMoM2b}, includes over 10,000 singleton pregnancies of nulliparous mothers. Each pregnancy has details from up to four visits: V1-V3 represent each trimester, while V4 captures delivery. The primary focus was to better understand adverse pregnancy outcomes (APOs), particularly PTB, defined as delivery before 37 weeks and affecting around 10\% of pregnancies. Predicting PTB is challenging due to its imbalanced nature and lack of prior birth indicators for nulliparous mothers. The study gathered over 4600 features per pregnancy.

We selected PTB versus Full Term Birth (FTB) as class labels. The dataset includes 7,923 FTB and 780 PTB cases, reflecting an imbalance. After preprocessing, as outlined by \citep{goretsky2021datanumom2b}, and incorporating Layer 2, Visit 3 features (360 features), each patient is represented as an image for further analysis.
\subsubsection{Madelon}
% Madelon ~\citep{guyon2006feature} is a synthetic dataset created for the feature selection challenge of the Advances in Neural Information Processing Systems (NIPS 2003). It presents a binary classification task involving continuous input features. The unique aspect of this challenge lies in its multivariate nature, posing a highly non-linear problem.

% This dataset comprises samples, each featuring 500 attributes. Initially balanced, the dataset was intentionally made imbalanced in the first phase of the experiment. This imbalance was introduced through the random removal of data samples, resulting in a distribution of 10 percent for class one and 90 percent for class zero. 
Madelon~\citep{guyon2006feature} is a synthetic dataset created for the NIPS 2003 feature selection challenge. It involves a binary classification task with continuous input features and a multivariate, highly non-linear nature. The dataset contains samples with 500 attributes. Initially balanced, it was made imbalanced in the experiment by randomly removing samples, resulting in a 10\% class one and 90\% class zero distribution.

% \begin{table*}[h!]
%   \centering
%   \small
%   \begin{tabular}{|c|c|c|c|}
%     \hline
%     \textbf {Method} 
     
%     & \multicolumn{1}{|p{2cm}|}{\centering \textbf{FTB} \\ \textbf{vs rest}} 
%     & \multicolumn{1}{|p{2cm}|}{\centering \textbf{sPTB} \\ \textbf{vs rest}}
%     & \multicolumn{1}{|p{2cm}|}{\centering \textbf{iPTB} \\ \textbf{vs rest}}
%      \\
%     \hline
%      CNN w/o augmentation & 0.7085 ± 0.0186 &   0.6694 ± 0.0299& 0.7671 ± 0.0415
%   \\
%     \hline
%     CNN + augmentation & 0.6958 ± 0.0256 & 0.6615 ± 0.0330
%  & 0.7542 ± 0.0372\\
    
%  \hline
%     XGBoost w/o augmentation & 0.7406 ± 0.0185
%  & 0.7078 ± 0.0032 & 0.8009 ± 0.0396\\
%  \hline
%     XGBoost + augmentation & \textbf{0.7496 ± 0.0177}
%  & \textbf{0.7144 ± 0.0029} & \textbf{0.8099 ± 0.0355}\\
%     \hline
    
%   \end{tabular}
%   \caption{AUC scores for Experiment 3 on nuMoM2b dataset}
%   \label{table:AUC_Subgroup}
% \end{table*}

\subsubsection{Myocardial Infarction complications}
% The Myocardial infarction complications dataset~\citep{misc_myocardial_infarction_complications_579} comprising 1700 records from Krasnoyarsk Interdistrict Clinical Hospital, Russia, was meticulously compiled. This dataset, released in August 2020 and accessed from the UCI Machine Learning Repository, serves as a crucial resource for this investigation. It encompasses a comprehensive set of 124 attributes. Of these, 111 attributes provide insights into patients' demographics, medical history, complications during hospital admission, ECG results, and subsequent clinical interventions. The remaining 12 columns delineate various complications observed across four distinct time stages. For the scope of this study, we selected Myocardial Rupture as the target of interest within the dataset, considering its imbalance.
The Myocardial Infarction Complications dataset~\citep{misc_myocardial_infarction_complications_579} contains 1700 records from Krasnoyarsk Interdistrict Clinical Hospital, Russia, released in August 2020 via the UCI Machine Learning Repository. It includes 124 attributes, with 111 detailing patient demographics, medical history, hospital complications, ECG results, and clinical interventions, while 12 describe complications across four time stages. This study focuses on Myocardial Rupture, selected for its class imbalance.

\subsubsection{DARWIN}

% The DARWIN dataset ~\citep{cilia2022darwin} includes handwriting data from 174 participants—89 with Alzheimer’s and 85 healthy individuals—collected using a graphic tablet during 25 handwriting tasks. The dataset captures fine motor skill impairments indicative of neurodegenerative diseases. The tasks and extracted features help identify differences in handwriting dynamics, providing a valuable resource for researchers aiming to develop machine learning models to diagnose Alzheimer's disease more accurately. This dataset contains samples, each with 450 attributes. Initially balanced, it is deliberately made imbalanced during the first phase of the experiment.
The DARWIN dataset~\citep{cilia2022darwin} comprises handwriting data from 174 participants (89 with Alzheimer’s and 85 healthy) collected via a graphic tablet across 25 tasks. It captures motor skill impairments indicative of neurodegenerative diseases. With 450 attributes per sample, the dataset aids in developing machine learning models for Alzheimer’s diagnosis. Initially balanced, it was deliberately imbalanced in the first phase of the experiment.

\subsection{Method}
% When assessing our {\it RIGA} approach, we follow a process that involves distinct training phases. Initially, we utilize a separate training folds for the image transformation technique and training generative models. Moreover, we performed all experiments using a 5-fold cross-validation approach, starting from the initial data transformation stage to the final classification stage.
When evaluating our \textit{RIGA} approach, we use separate training folds for image transformation and generative models, with all experiments conducted using 5-fold cross-validation from data transformation to final classification.

\subsubsection{Data Transformation}
Employing the DeepInsight method for transforming tabular data into image data involved specifying the image size as 28x28 to encompass all features and facilitate subsequent use with generative models. Additionally, we utilized t-SNE as a dimensionality reduction technique for this transformation.The process began with normalizing the data, followed by feeding it into the method to obtain 28x28 images from the original tabular data.

\subsubsection{Augmentation}
During the training phase, the training dataset is utilized for both image transformation and generative models training.

% \begin{enumerate}
% \item \textbf{cGAN}: The cGAN is trained for 50 epochs. The generator takes a 100-dimensional random noise vector and corresponding labels as input. In evaluation, the cGAN generates synthetic data, which is added to the training set and used to train our CNN or other machine learning models.

% \item \textbf{VQ-VAE}: The VQ-VAE model is trained for 50 epochs. The encoder quantizes input images using 128 embeddings, and the decoder reconstructs them from quantized latents. After training, VQ-VAE uses PixelCNN to generate prior distributions, which are decoded into synthetic images. These generated samples are added to the training dataset to enhance downstream models like CNNs or other machine learning algorithms.

% \item \textbf{VQ-GAN}: 
% The VQ-GAN model is trained for 50 epochs. The encoder quantizes input images with 128 embeddings, and a transformer captures long-range dependencies. The decoder reconstructs images, while a discriminator improves quality. During evaluation, synthetic data is generated from random latent vectors and integrated into the training dataset for further analysis.
% \end{enumerate}
The cGAN is trained for 50 epochs, with the generator taking a 100-dimensional random noise vector and labels as input. During evaluation, it generates synthetic data, which is added to the training set to improve CNN or other machine learning models. The VQ-VAE model, also trained for 50 epochs, uses an encoder to quantize images with 128 embeddings and a decoder to reconstruct them. After training, PixelCNN generates priors that are decoded into synthetic images, which are added to the training set for enhancing CNNs or other models. Similarly, the VQ-GAN model, trained for 50 epochs, employs an encoder with 128 embeddings, a transformer for long-range dependencies, and a decoder refined by a discriminator. Synthetic data is generated from random latent vectors during evaluation and incorporated into the training dataset for further analysis.

Figure \ref{fig:real_fake_images} in the appendix describes snapshots of real and synthetic images generated by the generative models trained for the four transformed datasets.

\subsubsection{Classification}

In the classification phase, we initially generated a set of images from the underrepresented class to create a balanced dataset. At this phase, we had two alternatives: (a) Employing a CNN with the entire image dataset; (b) Employing the inverse transform to revert the data to tabular form and utilizing traditional machine learning techniques. For conventional machine learning, XGBoost was employed.

Through initial experiments with augmented images, we selected the CNN architecture by performing hyperparameter tuning using grid search. This involved optimizing batch size, the number of layers, and the number of neurons to identify the best configuration.

We use AUC (Area Under the Curve) as the performance metric because it is robust to class imbalance and independent of threshold selection. Unlike accuracy, AUC remains reliable regardless of class ratios and evaluates model performance across all thresholds, providing a comprehensive view of discrimination power. This made AUC ideal for assessing our models on imbalanced data.

\subsubsection{Bayesian Network Learning}

% The experiments to be conducted on the previously discussed U2 pipeline were based on fundamental comparison and observation noted in the change in the scoring function. Initially, we analyze a dataset characterized by imbalanced class distributions. Following the U2 process on the original dataset, we obtained a final BIC score that reflected the model's performance. With the RIGA pipeline, the augmented dataset has a more balanced class distribution. After the dataset is balanced, we perform this evaluation phase on the augmented data. Apart from observing the change in the scoring function, we also visualized the learnt Bayesian Network with the help of Markov Blanket. Once the training phase of the Bayesian network is completed, the Markov blanket assists us in working with any variable and predicting relevant information. As the label distribution of the dataset improves, the representation of the underlying relationships within the data becomes clearer, which can lead to a more accurate and reliable representation of the Markov Blanket in a Bayesian Network.
The experiments conducted on the U2 pipeline focused on comparing the scoring function changes. Initially, we analyzed a dataset with imbalanced class distributions. The U2 process on the original dataset produced a BIC score reflecting the model's performance. Using the RIGA pipeline, the augmented dataset achieved a more balanced distribution, enabling improved evaluation. The scoring function change was analyzed alongside the visualization of the learned Bayesian Network using the Markov Blanket. Once trained, the Markov Blanket aids in predicting relevant information for any variable. As class balance improves, the representation of underlying data relationships becomes clearer, enhancing the reliability of the Bayesian Network.

\begin{table*}[!ht]
  \centering
  \small
  \begin{tabular}{|c|l|c|c|c|c|c|c|c||}
    \hline
    \textbf {Method} 
    & \multicolumn{1}{|p{2cm}|}{\centering \textbf{nuMoM2b}} 
    & \multicolumn{1}{|p{2cm}|}{\centering \textbf{Madelon}} 
    & \multicolumn{1}{|p{2cm}|}{\centering \textbf{MI}}
    & \multicolumn{1}{|p{2cm}|}{\centering \textbf{DARWIN}}\\
    \hline
    CNN w/o Augmentation & 0.7086 ± 0.0105 & 0.6072 ± 0.0110 & 0.6551 ± 0.0849 & 0.8152 ± 0033 \\
    \hline
    CNN \& cGAN & 0.7136 ± 0.0220 & 0.6162 ± 0.0168 & 0.6139 ± 0.0793 & 0.8324 ± 0027 \\
    \hline
    CNN \& VQVAE & 0.7309 ± 0.0140 & 0.6334 ± 0.0215 & 0.6398 ± 0.0832 & 0.8575 ± 0021 \\
    \hline
    CNN \& VQGAN & 0.7368 ± 0.115 & 0.6512 ± 0.0173 & 0.6032 ± 0.0856 & 0.8324 ± 0042 \\
    \hline
    XGBoost w/o Augmentation & 0.7384 ± 0.0114 & 0.6738 ± 0.0490 & 0.8554 ± 0.0695 & 0.9252 ± 0010  \\
    \hline
    XGBoost \& cGAN & 0.7566 ± 0.0126 & 0.6901 ± 0.0341 & 0.8565 ± 0.0693 & 0.9350 ± 0029 \\
    \hline
    XGBoost \& VQVAE & 0.7572 ± 0.0135 & 0.7066 ± 0.0345 & \textbf{0.8615  ± 0.0828} & \textbf{0.9586 ± 0.0046}  \\
    \hline
    XGBoost \& VQGAN & \textbf{0.7598 ± 0.0118} & \textbf{0.7355 ± 0.0505} & 0.8459  ± 0.0784 & 0.9439 ± 0.0047 \\
    \hline
    XGBoost \& ADASYN & 0.7162 ± 0.0060 & 0.6058 ± 0.0520 & 0.8342 ± 0.0990 & 0.9240 ± 0.0035 \\
    \hline
    XGBoost \& SMOTE & 0.7183 ± 0.0030 & 0.6187 ± 0.0335 & 0.8366 ± 0.0812 & 0.9240 ± 0.0045 \\
    \hline
    XGBoost \& CTGAN & 0.7442 ± 0.0138 & 0.6543 ± 0.0101 & 0.8321 ± 0.0937 & 0.9339 ± 0023 \\
    \hline
  \end{tabular}
  \caption{AUC scores for different methods across datasets}
  \label{table:AUC_by_method}
\end{table*}

%% file: results.tex
\section{Empirical Results}
\label{sec:results}

% In this section, we describe the  experiments conducted for each dataset to assess whether our pipeline can enhance the outcomes. For the outcomes across all datasets, we employed a 5-fold cross-validation spanning from the initial phase of image transformation, generative models training, image generation, to the concluding classification stage.

In this section, we describe the experiments conducted on each dataset to evaluate the pipeline's effectiveness. A 5-fold cross-validation was applied from image transformation and generative model training to image generation and classification.
% \anita{as discussed choose 2 datasets only and go deep, you can add the \% improvements to Table 1; finally the goodfellow citation is incomplete. Could you complete it? I can do another read Wednesday morning at 9am.}
\subsection{nuMoM2b Results}
In the nuMoM2b dataset, binary classification distinguishes PTB from FTB cases. ADASYN and SMOTE decreased XGBoost AUC, while CTGAN provided minimal improvement. For CNN classification, RIGA with cGAN, VQVAE, and VQGAN augmentation improved performance by 0.5\%, 2.23\%, and 2.82\%, respectively. After converting to tabular data, RIGA with cGAN, VQVAE, and VQGAN enhanced XGBoost results by 1.82\%, 1.88\%, and 2.14\%, respectively.

% XGBoost using VQGAN outperformed because it combines VQVAE's discrete latent space modeling with GAN's adversarial training, producing higher-quality synthetic images that better represent the minority class, enhancing classifier performance; VQVAE was slightly better than cGAN due to its effective data representation. 
% \anita{ this is where you should really get into the details of the data and why one approach does better than the other and why? The details are in the data and thenyou can take that and generalize so the readers can identify similar structure in their data and understand their problem, data and solution better as well. This is what gets a paper accepted.}
% \mahdi{changed}
The performance difference observed in using VQGAN with XGBoost over other methods, such as VQVAE and cGAN, can be attributed to the nature of the numom2b dataset and the task at hand. With 360 features and an imbalanced binary classification problem (7,923 FTB vs. 780 PTB cases), capturing the nuanced patterns of the minority class (PTB) is critical. VQGAN's ability to leverage VQVAE's discrete latent space modeling enhances the representation of complex features, allowing it to generate synthetic samples that better reflect the distribution and variability of PTB cases. This improves the classifier's ability to distinguish minority class instances. In contrast, VQVAE alone, while effective at data representation, lacks the adversarial training that refines the synthetic outputs for higher fidelity, which explains its slightly lower performance. Meanwhile, cGAN, which depends on a continuous latent space, might struggle to model the discrete and intricate feature distributions inherent in the data, leading to relatively poorer representation and classification performance. These insights highlight the importance of aligning data generation methods with the structural properties of the dataset to improve model performance, offering generalizable lessons for other imbalanced datasets.

\subsection{Madelon Results} \label{Madelon_Dataset}
% In the Madelon dataset, we deal with a binary classification problem involving synthetic data. The dataset comprises 2600 samples, each containing 500 features. To induce imbalance, we removed 1150 samples from class 1. Table \ref{table:AUC_by_method} presents the experimental outcomes for the Madelon dataset using various methods.

% For the Madelon dataset, balancing with synthetic data using ADASYN, SMOTE, and CTGAN all resulted in decreased AUC scores. RIGA approach with cGAN, VQVAE, and VQGAN, however, achieved improvements in AUC for CNN of 0.9\%, 2.62\%, and 4.4\%, respectively. When using XGBoost with a transformation back to tabular data, RIGA with cGAN showed a 1.63\% improvement in AUC, RIGA with VQVAE showed a 3.28\% improvement, and RIGA with VQGAN demonstrated a significant 6.17\% enhancement in AUC compared to XGBoost without augmentation.

% In this dataset, VQGAN performed best because it combines vector quantization with adversarial training, producing high-quality and diverse synthetic images that most effectively enhance classifier performance. VQVAE, lacking adversarial training, was second best. cGAN, while using adversarial training, doesn't utilize vector quantization, resulting in less effective data augmentation.
In the Madelon dataset, a binary classification task with 2600 samples and 500 features, imbalance was induced by removing 1150 class 1 samples. Table \ref{table:AUC_by_method} summarizes the results. ADASYN, SMOTE, and CTGAN reduced AUC scores. RIGA with cGAN, VQVAE, and VQGAN improved CNN AUC by 0.9\%, 2.62\%, and 4.4\%, respectively. For XGBoost with tabular transformations, RIGA with cGAN, VQVAE, and VQGAN enhanced AUC by 1.63\%, 3.28\%, and 6.17\%, respectively. VQGAN outperformed others by combining vector quantization with adversarial training for diverse, high-quality synthetic data. VQVAE, lacking adversarial training, ranked second, while cGAN, lacking vector quantization, was less effective.

\subsection{Myocardial Infarction Results}
In the Myocardial dataset, a binary classification problem, features with over 150 missing samples were excluded, and remaining samples with missing values were removed, resulting in 1024 samples and 94 features (1050 class zero, 24 class one). Table \ref{table:AUC_by_method} summarizes the results.Table \ref{table:AUC_by_method} summarizes the experimental results for the Myocardial dataset across different methods. ADASYN, SMOTE, and CTGAN reduced AUC scores, and CNN performed poorly, likely due to sparse feature representation in pixels. For XGBoost with RIGA, cGAN improved AUC by 0.1\%, VQVAE by 0.6\%, and VQGAN decreased AUC. These outcomes reflect the limited ability of models to capture meaningful structures due to sparse feature representation during transformation.
\subsection{DARWIN Results}
% \mahdi{changed}
In the DARWIN dataset, we address a binary classification problem involving 451 features, one of which was an ID column removed prior to analysis. To induce class imbalance, we employed the same sample removal strategy outlined in Section \ref{Madelon_Dataset}, ensuring consistency across datasets. The final dataset contains 174 samples, a notably small size, which poses challenges for training and evaluating machine learning models. Table \ref{table:AUC_by_method} presents the experimental results for the DARWIN dataset.

For this dataset, both ADASYN and SMOTE reduced AUC scores, highlighting their inability to generate synthetic samples that effectively represent the feature distributions in such a small dataset. CTGAN, by leveraging adversarial training, provided a modest improvement of 0.87\%, showcasing its ability to capture more meaningful feature distributions. RIGA, combined with cGAN, achieved a slightly higher improvement of 0.98\%, likely benefiting from its adversarial training tailored to specific transformations.

Interestingly, RIGA with VQVAE delivered the highest improvement, boosting AUC by 3.34\%. This can be attributed to VQVAE's simpler architecture, which is well-suited for small datasets like DARWIN, as it avoids the overfitting risks associated with more complex models like VQGAN. In comparison, RIGA with VQGAN yielded a 1.87\% improvement, falling short of VQVAE due to its complexity, which can struggle in scenarios with limited data. These results underscore the importance of selecting augmentation techniques that align with the size and structure of the dataset to maximize performance gains.

\subsection{Bayesian Network Learning}
% \begin{figure}[h]
%     \centering
%     \begin{tikzpicture}
%         % Nodes
%         \node (A) [font=\small] {Days of Vaginal Bleeding};
%         \node (B) [below=0.15cm of A, font=\small] {Cervical Length};
%         \node (C) [right=1.0cm of A, font=\small] {Outcome};

%         % Draw rectangles around each node
%         \draw[rounded corners] (A.north west) rectangle (A.south east);
%         \draw[rounded corners] (B.north west) rectangle (B.south east);
%         \draw[rounded corners] (C.north west) rectangle (C.south east);

%         % Edges
%         \draw[->] (A) -- (C);
%         \draw[->] (B) -- (C);
%     \end{tikzpicture}
%     \caption{Markov Blanket without RIGA}
%     \label{fig:MBwoRIGA}
% \end{figure}
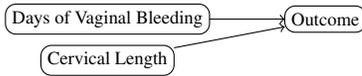
\begin{figure}[h]
    \centering
    \scriptsize % Adjust font size for the entire TikZ picture
    \begin{tikzpicture}
        % Nodes
        \node (A) [font=\scriptsize] {Days of Vaginal Bleeding};
        \node (B) [below=0.15cm of A, font=\scriptsize] {Cervical Length};
        \node (C) [right=1.0cm of A, font=\scriptsize] {Outcome};

        % Draw rectangles around each node
        \draw[rounded corners] (A.north west) rectangle (A.south east);
        \draw[rounded corners] (B.north west) rectangle (B.south east);
        \draw[rounded corners] (C.north west) rectangle (C.south east);

        % Edges
        \draw[->] (A) -- (C);
        \draw[->] (B) -- (C);
    \end{tikzpicture}
    \caption{Markov Blanket without RIGA}
    \label{fig:MBwoRIGA}
\end{figure}

% Second graph: D, E, F directing to C, each with its own rectangle
% \begin{figure}[h]
%     \centering
%     \begin{tikzpicture}
%         % Nodes
%         \node (D) [font=\small] {Tobacco products smoked (3rd visit)};
%         \node (E) [below=0.15cm of D, font=\small] {Tobacco products smoked (2nd visit)};
%         \node (F) [below=0.15cm of E, font=\small] {Chest abnormality};
%         \node (C) [right=1.0cm of D, font=\small] {Outcome};

%         % Draw rectangles around each node
%         \draw[rounded corners] (D.north west) rectangle (D.south east);
%         \draw[rounded corners] (E.north west) rectangle (E.south east);
%         \draw[rounded corners] (F.north west) rectangle (F.south east);
%         \draw[rounded corners] (C.north west) rectangle (C.south east);

%         % Edges
%         \draw[->] (D) -- (C);
%         \draw[->] (E) -- (C);
%         \draw[->, bend right=10] (F.east) to (C.west);
%     \end{tikzpicture}
%     \caption{Markov Blanket with RIGA}
%     \label{fig:MBwRIGA}
% \end{figure}
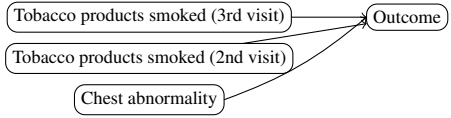
\begin{figure}[h]
    \centering
    \scriptsize % Adjust font size for the entire TikZ picture
    \begin{tikzpicture}
        % Nodes
        \node (D) [font=\scriptsize] {Tobacco products smoked (3rd visit)};
        \node (E) [below=0.15cm of D, font=\scriptsize] {Tobacco products smoked (2nd visit)};
        \node (F) [below=0.15cm of E, font=\scriptsize] {Chest abnormality};
        \node (C) [right=1.0cm of D, font=\scriptsize] {Outcome};

        % Draw rectangles around each node
        \draw[rounded corners] (D.north west) rectangle (D.south east);
        \draw[rounded corners] (E.north west) rectangle (E.south east);
        \draw[rounded corners] (F.north west) rectangle (F.south east);
        \draw[rounded corners] (C.north west) rectangle (C.south east);

        % Edges
        \draw[->] (D) -- (C);
        \draw[->] (E) -- (C);
        \draw[->, bend right=10] (F.east) to (C.west);
    \end{tikzpicture}
    \caption{Markov Blanket with RIGA}
    \label{fig:MBwRIGA}
\end{figure}

As mentioned earlier, we compare the effect of RIGA on our U2 pipeline. Initially, the BIC score for the model trained on the imbalanced dataset was {\it -4.618e5}. After applying the RIGA pipeline, which augmented the dataset for a more balanced class distribution, the BIC score improved to {\it -4.855e5}. This decrease indicates that the Bayesian network fitted the augmented data more effectively, enhancing the representation of underlying relationships. Visualization of the learned Bayesian Network through the Markov Blanket revealed expanded variable interactions. Initially, the Markov blanket of the target variable included two parent features, which increased to three after augmentation, as shown in Figures \ref{fig:MBwoRIGA} and \ref{fig:MBwRIGA}. This expansion highlights improved clarity in variable interactions as class balance improved. Notably, the absence of certain features in the improved Markov Blanket doesn’t imply their insignificance. These results underscore the importance of addressing class imbalances, enhancing both predictive capabilities and the understanding of data structure, thereby improving real-world applications.

%% file: summary.tex
\section{Summary of results}
\label{sec:discussion}

Based on our results, VQGAN within the RIGA pipeline performed better on the nuMoM2b and Madelon datasets, which have larger sample sizes (360 and 500 features). Conversely, VQVAE excelled on the myocardial infarction and DARWIN datasets, which have smaller sample sizes (1,024 and 174 samples) and fewer features (94 and 450). This indicates that VQGAN's complex architecture benefits larger datasets, while VQVAE's simpler design suits smaller datasets, reducing overfitting.
For instance, VQGAN effectively captures pixel-feature correlations in the nuMoM2b dataset but fails in the MI dataset, generating unrealistic single high-intensity pixel images, as shown in Figure \ref{fig:riga_vqgan} in the appendix.

% \begin{figure}[ht]
%     \centering
%     \includegraphics[width=0.4\textwidth]{images/Blank diagram.png} % Adjust the width as needed
%     \caption{Real and Synthetic Images Generated by RIGA-VQGAN for nuMoM2b and MI Datasets}
%     \label{fig:fake_real_mi_nu}
% \end{figure}

%% file: conclusions.tex
\section{Conclusions and Future Work}
\label{sec:conclusions}
This work introduces RIGA, a novel pipeline addressing uncertainty and class imbalance in high-stakes AI applications like healthcare. By transforming tabular data into images, RIGA leverages generative models—cGAN, VQVAE, and VQGAN—to create balanced synthetic samples, improving classification performance. RIGA’s inverse transformation enables generated images to be reverted back to their original tabular form, allowing seamless integration into downstream tasks.
Experiments show RIGA enhances traditional classifiers like XGBoost, with VQGAN improving performance by 6.17\% on Madelon and 2.14\% on nuMoM2b, while VQVAE achieved 3.34\% improvement on DARWIN. RIGA also strengthens Bayesian structure learning, enhancing robustness in imbalanced health datasets. Future work will incorporate an Error Feedback Loop to refine generative models and explore data augmentation’s role in Bayesian learning to uncover hidden feature correlations.

%% file: ack.tex
\section{Acknowledgements}
\label{sec:ack}

Research reported in this publication was supported
by the National Library Of Medicine of the National
Institutes of Health under Award Number
R01LM013327 and a  PSC-CUNY 2023 . The content is solely the responsibility
of the authors and does not necessarily represent
the official views of the National Institutes of Health.

%% file: appendix.tex
% \clearpage
% \section{Appendix}

% \begin{figure}[ht]
%     \centering
%     % First figure in a minipage
%     \begin{minipage}[t]{0.6\textwidth}
%         \centering
%         \includegraphics[width=\textwidth]{images/real_fake_images (1).png}
%         \captionsetup{justification=centering}
%         \caption{\centering Real and Fake Images of four datasets}
%         \label{fig:real+fake}
%     \end{minipage}
%     \hspace{0.05\textwidth} % Horizontal space between figures
%     % Second figure in a minipage
%     \begin{minipage}[t]{0.3\textwidth}
%         \centering
%         \includegraphics[width=\textwidth]{images/Blank diagram.png}
%         \caption{Real and Synthetic Images Generated by RIGA-VQGAN for nuMoM2b and MI Datasets}
%         \label{fig:fake_real_mi_nu}
%     \end{minipage}
% \end{figure}

% \clearpage
% \section{Appendix}

% \begin{figure}[ht]
%     \centering
%     \includegraphics[width=\linewidth]{images/real_fake_images (1).png}
%     \caption{Real and Fake Images of four datasets.}
%     \label{fig:real_fake}
% \end{figure}

% \begin{figure}[ht]
%     \centering
%     \includegraphics[width=\linewidth]{images/Blank diagram.png}
%     \caption{Real and Synthetic Images Generated by RIGA-VQGAN for nuMoM2b and MI Datasets.}
%     \label{fig:fake_real_mi_nu}
% \end{figure}

\onecolumn
\section{Appendix}

\begin{figure}[ht] \label{real}
    \centering
    \includegraphics[width=0.5\linewidth]{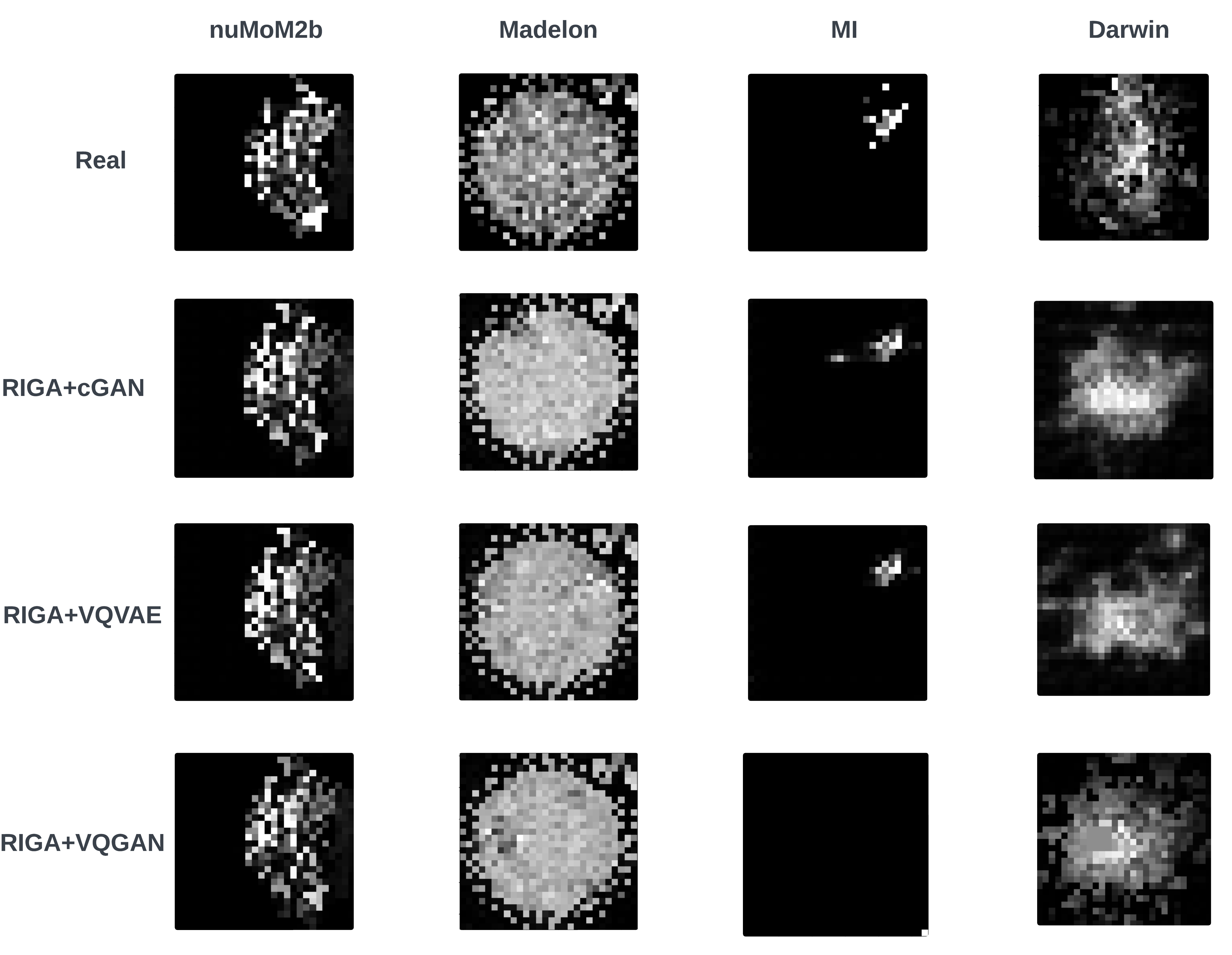}
    \caption{Real and Fake Images of four datasets.}
    \label{fig:real_fake_images}
\end{figure}

\begin{figure}[ht]
    \centering
    \includegraphics[width=0.5\linewidth]{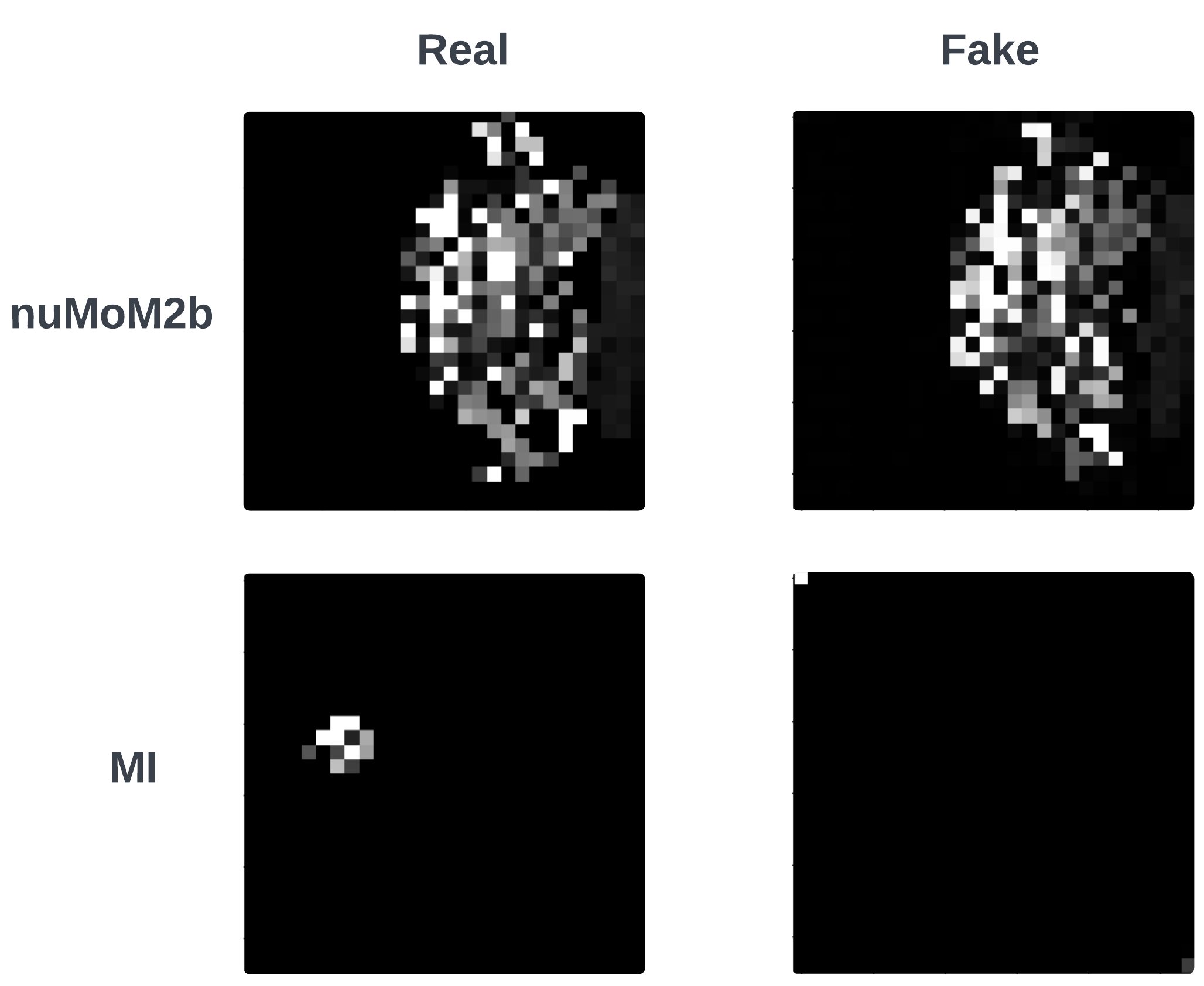}
    \caption{Real and Synthetic Images Generated by RIGA-VQGAN for nuMoM2b and MI Datasets.}
    \label{fig:riga_vqgan}

\end{figure}

%% file: anonymous-submission-latex-2025.bbl
\begin{thebibliography}{28}
\providecommand{\natexlab}[1]{#1}

\bibitem[{Aggarwal(2018)}]{aggarwal2018neural}
Aggarwal, C.~C. 2018.
\newblock \emph{Neural Networks and Deep Learning}, volume~10.
\newblock Springer.

\bibitem[{Andresini et~al.(2021)Andresini, Appice, Rose, and Malerba}]{ANDRESINI2021108}
Andresini, G.; Appice, A.; Rose, L.~D.; and Malerba, D. 2021.
\newblock GAN augmentation to deal with imbalance in imaging-based intrusion detection.
\newblock \emph{Future Generation Computer Systems}, 123: 108--127.

\bibitem[{Chawla et~al.(2002)Chawla, Bowyer, Hall, and Kegelmeyer}]{chawla2002smote}
Chawla, N.~V.; Bowyer, K.~W.; Hall, L.~O.; and Kegelmeyer, W.~P. 2002.
\newblock SMOTE: synthetic minority over-sampling technique.
\newblock \emph{Journal of artificial intelligence research}, 16: 321--357.

\bibitem[{Cilia et~al.(2022)Cilia, De~Gregorio, De~Stefano, Fontanella, Marcelli, and Parziale}]{cilia2022darwin}
Cilia, N.~D.; De~Gregorio, G.; De~Stefano, C.; Fontanella, F.; Marcelli, A.; and Parziale, A. 2022.
\newblock Diagnosing Alzheimer’s disease from on-line handwriting: A novel dataset and performance benchmarking.
\newblock \emph{Engineering Applications of Artificial Intelligence}, 111: 104822.

\bibitem[{Esser, Rombach, and Ommer(2021)}]{esser2021taming}
Esser, P.; Rombach, R.; and Ommer, B. 2021.
\newblock Taming transformers for high-resolution image synthesis.
\newblock In \emph{Proceedings of the IEEE/CVF Conference on Computer Vision and Pattern Recognition}, 12873--12883.

\bibitem[{Golovenkin et~al.(2020)Golovenkin, Shulman, Rossiev, Shesternya, Nikulina, Orlova, and Voino-Yasenetsky}]{misc_myocardial_infarction_complications_579}
Golovenkin, S.; Shulman, V.; Rossiev, D.; Shesternya, P.; Nikulina, S.; Orlova, Y.; and Voino-Yasenetsky, V. 2020.
\newblock {Myocardial infarction complications}.
\newblock UCI Machine Learning Repository.
\newblock {DOI}: https://doi.org/10.24432/C53P5M.

\bibitem[{Goodfellow et~al.(2014)Goodfellow, Pouget-Abadie, Mirza, Xu, Warde-Farley, Ozair, Courville, and Bengio}]{goodfellow2014generative}
Goodfellow, I.; Pouget-Abadie, J.; Mirza, M.; Xu, B.; Warde-Farley, D.; Ozair, S.; Courville, A.; and Bengio, Y. 2014.
\newblock Generative adversarial nets.
\newblock In \emph{Advances in neural information processing systems}, volume~27.

\bibitem[{Goretsky et~al.(2021)Goretsky, Dmitrienko, Tang, Lari, Kunhardt, Khan, and Gyamfi-Bannerman}]{goretsky2021datanumom2b}
Goretsky, A.; Dmitrienko, A.; Tang, I.; Lari, N.; Kunhardt, O.; Khan, R.~R.; and Gyamfi-Bannerman, C. 2021.
\newblock Data Preparation of the nuMoM2b Dataset.
\newblock \emph{medRxiv}.
\newblock Preprint.

\bibitem[{Guyon et~al.(2006)Guyon, Gunn, Nikravesh, and Zadeh}]{guyon2006feature}
Guyon, I.; Gunn, S.; Nikravesh, M.; and Zadeh, L.~A. 2006.
\newblock \emph{Feature Extraction: Foundations and Applications}.
\newblock Springer Science \& Business Media.

\bibitem[{Haas et~al.(2015)Haas, Parker, Wing, Parry, Grobman, Mercer, Simhan, Hoffman, Silver, Wadhwa, Iams, Koch, Caritis, Wapner, Esplin, Elovitz, Foroud, Peaceman, Saade, Willinger, and Reddy}]{Haas2015nuMoM2b}
Haas, D.~M.; Parker, C.~B.; Wing, D.~A.; Parry, S.; Grobman, W.~A.; Mercer, B.~M.; Simhan, H.~N.; Hoffman, M.~K.; Silver, R.~M.; Wadhwa, P.; Iams, J.~D.; Koch, M.~A.; Caritis, S.~N.; Wapner, R.~J.; Esplin, M.~S.; Elovitz, M.~A.; Foroud, T.; Peaceman, A.~M.; Saade, G.~R.; Willinger, M.; and Reddy, U.~M. 2015.
\newblock A description of the methods of the {Nulliparous} {Pregnancy} {Outcomes} {Study}: monitoring mothers-to-be ({nuMoM2b}).
\newblock \emph{American Journal of Obstetrics and Gynecology}, 212(4): 539.e1--539.e24.

\bibitem[{He et~al.(2008)He, Bai, Garcia, and Li}]{adasyn}
He, H.; Bai, Y.; Garcia, E.~A.; and Li, S. 2008.
\newblock ADASYN: Adaptive synthetic sampling approach for imbalanced learning.

\bibitem[{Huang et~al.(2018)Huang, Woodruff, Baer, Bangia, August, Jellife-Palowski, ..., and Sirota}]{huang2018investigation}
Huang, H.; Woodruff, T.~J.; Baer, R.~J.; Bangia, K.; August, L.~M.; Jellife-Palowski, L.~L.; ...; and Sirota, M. 2018.
\newblock Investigation of association between environmental and socioeconomic factors and preterm birth in California.
\newblock \emph{Environment international}, 121: 1066--1078.

\bibitem[{Iams et~al.(2001)Iams, Goldenberg, Mercer, Moawad, Meis, Das, and Roberts}]{iams2001preterm}
Iams, J.~D.; Goldenberg, R.~L.; Mercer, B.~M.; Moawad, A.~H.; Meis, P.~J.; Das, A.~F.; and Roberts, J.~H. 2001.
\newblock The preterm prediction study: Can low-risk women destined for spontaneous preterm birth be identified?
\newblock \emph{American journal of obstetrics and gynecology}, 184(4): 652--655.

\bibitem[{Koller and Friedman(2009)}]{Koller2009}
Koller, D.; and Friedman, N. 2009.
\newblock \emph{Probabilistic Graphical Models: Principles and Techniques}.
\newblock Cambridge Massachusetts: The MIT Press.

\bibitem[{Koullali et~al.(2016)Koullali, Oudijk, Nijman, Mol, and Pajkrt}]{koullali2016risk}
Koullali, B.; Oudijk, M.~A.; Nijman, T.~A.; Mol, B. W.~J.; and Pajkrt, E. 2016.
\newblock Risk assessment and management to prevent preterm birth.
\newblock \emph{Seminars in Fetal and Neonatal Medicine}, 21(2): 80--88.

\bibitem[{Maharana, Mondal, and Nemade(2022)}]{maharana2022review}
Maharana, K.; Mondal, S.; and Nemade, B. 2022.
\newblock A Review: Data Pre-processing and Data Augmentation Techniques.
\newblock \emph{Global Transitions Proceedings}, 3(1): 91--99.

\bibitem[{Mallia(2023)}]{Mallia2023}
Mallia, D. 2023.
\newblock \emph{Towards an unsupervised Bayesian network pipeline for explainable prediction, decision making and discovery}.
\newblock Ph.D. thesis, City University of New York %(CUNY).

\bibitem[{Manuck(2017)}]{manuck2017racial}
Manuck, T.~A. 2017.
\newblock Racial and ethnic differences in preterm birth: a complex, multifactorial problem.
\newblock \emph{Seminars in perinatology}, 41(8): 511--518.

\bibitem[{Mikołajczyk and Grochowski(2018)}]{mikolajczyk2018data}
Mikołajczyk, A.; and Grochowski, M. 2018.
\newblock Data Augmentation for Improving Deep Learning in Image Classification Problem.
\newblock In \emph{2018 International Interdisciplinary PhD Workshop (IIPhDW)}, 117--122. IEEE.

\bibitem[{Mirza and Osindero(2014)}]{cgan}
Mirza, M.; and Osindero, S. 2014.
\newblock Conditional Generative Adversarial Nets.
\newblock \emph{CoRR}, abs/1411.1784.

\bibitem[{Mumuni and Mumuni(2022)}]{mumuni2022data}
Mumuni, A.; and Mumuni, F. 2022.
\newblock Data augmentation: A comprehensive survey of modern approaches.
\newblock \emph{Array}, 100258.

\bibitem[{Nanni et~al.(2021)Nanni, Paci, Brahnam, and Lumini}]{nanni2021comparison}
Nanni, L.; Paci, M.; Brahnam, S.; and Lumini, A. 2021.
\newblock Comparison of Different Image Data Augmentation Approaches.
\newblock \emph{Journal of Imaging}, 7(12): 254.

\bibitem[{Purisch and Gyamfi-Bannerman(2017)}]{purisch2017epidemiology}
Purisch, S.~E.; and Gyamfi-Bannerman, C. 2017.
\newblock Epidemiology of preterm birth.
\newblock \emph{Seminars in perinatology}, 41(7): 387--391.

\bibitem[{Sharma et~al.(2019)Sharma, Vans, Shigemizu, Boroevich, and Tsunoda}]{sharma2019deepinsight}
Sharma, A.; Vans, E.; Shigemizu, D.; Boroevich, K.~A.; and Tsunoda, T. 2019.
\newblock DeepInsight: A methodology to transform a non-image data to an image for convolution neural network architecture.
\newblock \emph{Scientific Reports}, 9(1): 11399.

\bibitem[{Van Den~Oord and Vinyals(2017)}]{vandenoord2017neural}
Van Den~Oord, A.; and Vinyals, O. 2017.
\newblock Neural discrete representation learning.
\newblock \emph{Advances in Neural Information Processing Systems}, 30.

\bibitem[{Xu et~al.(2019)Xu, Skoularidou, Cuesta-Infante, and Veeramachaneni}]{ctgan}
Xu, L.; Skoularidou, M.; Cuesta-Infante, A.; and Veeramachaneni, K. 2019.
\newblock Modeling Tabular data using Conditional GAN.

\bibitem[{Xu et~al.(2023)Xu, Yoon, Fuentes, and Park}]{image_aug_survey}
Xu, M.; Yoon, S.; Fuentes, A.; and Park, D.~S. 2023.
\newblock A Comprehensive Survey of Image Augmentation Techniques for Deep Learning.
\newblock \emph{Pattern Recognition}, 137: 109347.

\bibitem[{Zhu et~al.(2021)Zhu, Brettin, Xia et~al.}]{Zhu2021}
Zhu, Y.; Brettin, T.; Xia, F.; et~al. 2021.
\newblock Converting tabular data into images for deep learning with convolutional neural networks.
\newblock \emph{Scientific Reports}, 11: 11325.

\end{thebibliography}
